\documentclass[a4paper,twoside]{article}

\usepackage[dvipdfmx]{graphicx}
\usepackage{epsfig}
\usepackage{subcaption}
\usepackage{calc}
\usepackage{amssymb}
\usepackage{amstext}
\usepackage{amsmath}
\usepackage{amsthm}
\usepackage{multicol}
\usepackage{pslatex}
\usepackage{apalike}
\usepackage[bottom]{footmisc}
\usepackage{SCITEPRESS}     

\usepackage{bm}
\usepackage{multirow}
\usepackage{wrapfig}
\usepackage{algorithm,algorithmic}

\usepackage{xcolor}         
\newif\ifdraft
\drafttrue

\ifdraft
	\newcommand{\ToDo}[1]{\textcolor{red}{#1}}
\else
	\newcommand{\ToDo}[1]{}
\fi
\newcommand{\ours}{$\mathrm{M^{2}AT}$}

\begin{document}

\title{Masking and Mixing Adversarial Training}


\ifdraft
    \author{\authorname{Hiroki Adachi\sup{1},
    Tsubasa Hirakawa\sup{1},
    Takayoshi Yamashita\sup{1},
    Hironobu Fujiyoshi\sup{1},
    Yasunori Ishii\sup{2},
    Kazuki Kozuka\sup{2}}
    \affiliation{\sup{1}Chubu University, 1200 Matsumoto-cho, Kasugai, Aichi, Japan}
    \affiliation{\sup{2}Panasonic Holdings Corporation, Japan}
    \email{\{ha618, hirakawa\}@mprg.cs.chubu.ac.jp,
    \{takayoshi, fujiyoshi\}@isc.chubu.ac.jp,\\
    \{ishii.yasunori, kozuka.kazuki\}@jp.panasonic.com
    }
    }
\else
    \author{Anonymized for review}
\fi


\abstract{While convolutional neural networks (CNNs) have achieved excellent performances in various computer vision tasks, they often misclassify with malicious samples, a.k.a. adversarial examples.
  Adversarial training is a popular and straightforward technique to defend against the threat of adversarial examples.
  Unfortunately, CNNs must sacrifice the accuracy of standard samples to improve robustness against adversarial examples when adversarial training is used.
  In this work, we propose Masking and Mixing Adversarial Training~(\ours) to mitigate the trade-off between accuracy and robustness.
  We focus on creating diverse adversarial examples during training.
  Specifically, our approach consists of two processes: 1) \textbf{masking} a perturbation with a binary mask and 2) \textbf{mixing} two partially perturbed images.
  Experimental results on CIFAR-10 dataset demonstrate that our method achieves better robustness against several adversarial attacks than previous methods.
  }

\onecolumn \maketitle \normalsize \setcounter{footnote}{0} \vfill

\section{\uppercase{Introduction}} \label{sec:introduction}

In computer vision, deep convolutional neural networks (CNNs) have achieved excellent performances for various tasks such as image classification~\cite{He_2016}, image generation~\cite{Brock_2019}, object detection~\cite{Redmon_2016}, and semantic segmentation~\cite{Long_2015}.
To achieve these excellent performances, an enormous number of training samples or an increased diversity of samples using data augmentation~\cite{Zhang_2018}\cite{Yun_2019}\cite{Qin_2020} is required.
In this way, CNNs become robust to naturally occurring noise, e.g., rotation or translation and changes to the lighting environment.
However, an image with malicious perturbation~\cite{Szegedy_2014}, a.k.a. adversarial examples, induces misclassification with high confidence for CNNs.
This perturbation in adversarial examples is imperceptible to humans because of slightly.
Adversarial examples influence not only recognition and classification but also semantic segmentation~\cite{Xie_2017}, object detection, and depth estimation~\cite{Yamanaka_2020}.
As this phenomenon leads to security concerns in CNNs-based AI systems (e.g., those for autonomous driving or medical diagnosis), there are various methods to defend against adversarial attack~\cite{Samangouei_2018}\cite{Meng_2017}\cite{Shafahi_2019}\cite{Wong_2020}\cite{Goodfellow_2015}\cite{Lee_2020}.

Among them, adversarial training~\cite{Madry_2018} is the most popular and effective to improve the vulnerability of CNNs.
Adversarial training trains adversarial examples generated by projected gradient descent~\cite{Madry_2018}.
While CNNs become robust against such adversarial perturbations with adversarial training, they degrades the classification performance for benign samples.
This trade-off between accuracy and robustness has been demonstrated both theoretically and empirically from various aspects by many researchers.

Schmidt \textit{et al.}~\cite{Schmidt_2018} theoretically proved that adversarial training requires vaster and more complex data than the standard training for obtaining robustness.
Yin \textit{et al.}~\cite{Yin_2019} investigated the trade-off issue by performing spectral analysis of clean samples, adversarial examples, and samples to which data augmentation (e.g., Gaussian noise or fog) had been applied and showed that the trade-off occurs because adversarial examples involve higher frequency components than clean samples do.
Tsipras \textit{et al.}~\cite{Tsipras_2019} demonstrated through various experiments that adversarial training and standard training capture different features.
We feel this is probably due to the frequency bandwidth, as suggested by Yin \textit{et al.}.
Lee \textit{et al.}~\cite{Lee_2020} investigated Adversarial Feature Overfitting (AFO), which involves model parameters optimized to an unexpected direction, and proposed Adversarial Vertex mixup (AVmixup) as a strategy for avoiding AFO.

To summarize the above works, the trade-off stems from overfitting, which is caused by adversarial training with a limited coverage dataset.
With this issue in mind, we propose a method for computing adversarial examples during adversarial training, which can obtain excellent robustness while maintaining standard accuracy.
AVmixup extended the coverage of the training manifold by using label smoothing and interpolation between clean samples and virtual perturbations, thus preserving overfitting.
Although the authors did not prove it experimentally, \cite{Guo_2019} showed that interpolated data can be mapped to another manifold.
While AVmixup can achieve an excellent performance, perturbed image variation for adversarial training remains limited.
To train with rich variation of perturbed images, we propose Masking and Mixing Adversarial Training (\ours), which mitigates the trade-off by mixing unequal magnitudes of perturbation in a sample.
Our method consists following two processes:
1) masking the perturbation with a binary mask, which is defined such that the perturbation is located inside/outside the rectangle, and
2) mixing two partially perturbed images with an arbitrary mixing rate sampled from a beta distribution.
In summary, our work makes the following contributions:
\begin{itemize}
  \item We propose a powerful defense method that can mitigate the gap between accuracy and robustness by using adversarial examples with richer variation than prior works during training.
  \item We demonstrate through experiments that our method achieves state-of-the-art robustness on CIFAR-10 and discuss the interesting phenomenon observed in the experiments.
\end{itemize}

\section{\uppercase{Preliminaries}}

\textbf{Notations.}
We conduct a training dataset $\mathcal{D} = \{\bm{x}_{i}, y_{i}\}_{i}^{n}$ where $\bm{x}_{i}\in \mathbb{R}^{c\times h\times w}$ is an image and $y_{i}\in \mathcal{Y}=\{0, 1, \dots, K-1\}$ is the ground truth to $\bm{x}_{i}$.
We denote a model $f:\mathbb{R}^{c\times h\times w}\rightarrow \mathbb{R}^{K}$ parameterized by $\bm{\theta}$, and a loss function $L(f_{\bm{\theta}}(\bm{x}_{i}), y_{i})$ is a cross entropy loss as follow:
\begin{eqnarray}
    \mathcal{L}(f_{\bm{\theta}}(\bm{x}_{i}), y_{i}) &=& -\log \sigma_{y_{i}}(f_{\theta}(\bm{x}_{i}))\\
    &=& -\log p_{y_{i}}(\bm{x}_{i})),
\end{eqnarray}
where $\sigma:\mathbb{R}^{K}\rightarrow [0,1]^{K}$ denotes a softmax function, and both $\sigma_{y_{i}}$ and $p_{y_{i}}$ denote the true class probability.
The $\ell_{p}$ distance is written $\|\bm{x}_{i}\|_{p}$, where $\|\bm{x}_{i}\|_{p} = \left(\sum_{i=1}^{n}|\bm{x}_{i}|^{p}\right)^{\frac{1}{p}}$.

In the classification problem, because it is difficult to observe the data distribution clearly, we use empirical risk minimization:
\begin{equation}
  \min_{\bm{\theta}}\mathbb{E}_{(\bm{x}_i, y_i)\in \mathcal{D}}\left[\mathcal{L}(f_{\bm{\theta}}(\bm{x}_i), y_{i})\right],
\end{equation}

Unfortunately, while a model trained with above equation can classify to unknown data (i.e., validation or test data) somewhat correctly, it is quite vulnerable to adversarial examples.
To protect against this threat, adversarial training updates the model parameters $\bm{\theta}$ based on the defined adversarial examples in $\ell_{p}$-ball, such that the center is $\bm{x}$ and the radius is $\epsilon$:
\begin{equation}
\min_{\bm{\theta}}\mathbb{E}_{(\bm{x}_{i}, y_{i})\in \mathcal{D}}\left[
  \max_{\|\bm{\delta}_{i}\|_{p}\leq \epsilon}\mathcal{L}(f_{\bm{\theta}}(\bm{x}_{i}+\bm{\delta}_{i}), y_{i})
\right].
\end{equation}
For $\ell_{p}$-ball, we often use $\ell_{2}$ or $\ell_{\infty}$, and the perturbation defined with $\ell_{2}$ causes large alterations to the input image because $\|\cdot\|_{\infty}<\|\cdot\|_{2}$.
In this work, we aim to construct a model that is robust against the perturbation with $\ell_{\infty}$.
The perturbation $\bm{\delta}_{i}$ is not empirically sampled from a space existing on an infinite perturbation, but we create it based on gradients of the model, generally~\cite{Goodfellow_2015}\cite{Carlini_2017}\cite{Madry_2018}.

\textbf{Fast gradient sign method (FGSM)~\cite{Goodfellow_2015}:~~} FGSM provides the perturbation with single step by multiplying radius $\epsilon$ of $l_{p}$-ball by the unit vector extracted sign for gradients w.r.t. input data $\bm{x}_{i}$:
\begin{equation}
  \hat{\bm{x}}_{i} = \bm{x}_{i} + \epsilon\cdot\mathrm{sign}\left(
    \nabla_{\bm{x}_{i}}\mathcal{L}(f_{\bm{\theta}}(\bm{x}_{i}), y_{i})
  \right).
\end{equation}
FGSM can compute adversarial examples $\hat{\bm{x}}_{i}$ simply and fast.

\textbf{Projected gradient descent (PGD)~\cite{Madry_2018}:~~} Unlike FGSM with single step, PGD is multi-step attack using a step size $\alpha < \epsilon$, as
\begin{equation}
\hat{\bm{x}}^{(t+1)}_{i} = \Pi_{\mathcal{B}[\bm{x}_{i}^{(0)}]}\left(
  \hat{\bm{x}}^{(t)}_{i} + \alpha\cdot\mathrm{sign}\left(\nabla_{\hat{\bm{x}}^{(t)}_{i}}
  \mathcal{L}(f_{\bm{\theta}}(\hat{\bm{x}}_{i}^{(t)}), y_{i})\right)
\right),
\label{eq:pgd}
\end{equation}
where $t\in \mathbb{N}$, $\mathcal{B}[\bm{x}_{i}^{(0)}]:= \{\hat{\bm{x}}_{i}\in \mathcal{X}\mid \|\bm{x}^{(0)}_{i} - \hat{\bm{x}}_{i}\|_{\infty}\leq \epsilon\}$ in the input space $\mathcal{X}$, and $\Pi$ is a projection function that brings outliers into $\mathcal{B}[\bm{x}_{i}^{(0)}]$.
PGD can compute more severe perturbations than FGSM by demanding that the amount of movement to each pixel fits within $\ell_{p}$-ball with the projection function $\Pi$.

\section{\uppercase{Method}} 

In this section, we propose Masking and Mixing Adversarial Training~(\ours).
We aim to improve robustness against adversarial attacks by creating variety adversarial examples during training and training them.
In addition, we aim to maintain a standard classification accuracy to benign samples.

\subsection{Overview} \label{sec:subsection3-1}

Our method computes adversarial perturbations $\bm{\delta}\in \mathbb{R}^{c\times h \times w}$ from the model gradient by iterative arbitrary rounds, as shown in Eq.~(\ref{eq:pgd}).
Computed perturbations do not apply directly to samples but rather via our two processes.

First, we extract only part of the perturbation with the binary mask and apply it to the corresponding regions of the image.
Adversarial examples created with this process are called \textit{partially perturbed images} in this paper.
The details of how to define and add this binary mask to the adversarial perturbation are discussed in Section~\ref{sec:mask}.

Next, we create perturbed images by interpolating two samples, which are perturbed inside/outside, with an arbitrary probability.
Previous works directly add the computed perturbation to the entire image.
Meanwhile, adversarial examples created by our method inherent two type of magnitude of perturbation.
We define ground truth label to adversarial examples with a label smoothing.
Unlike the classical label smoothing~\cite{Szegedy_2016}, we use dynamic smoothing parameter sampled from the beta distribution during training.
The details of these processes are described in Section~\ref{sec:mix}.

\begin{figure*}[tb]
  \centering
  \includegraphics[width=.98\linewidth]{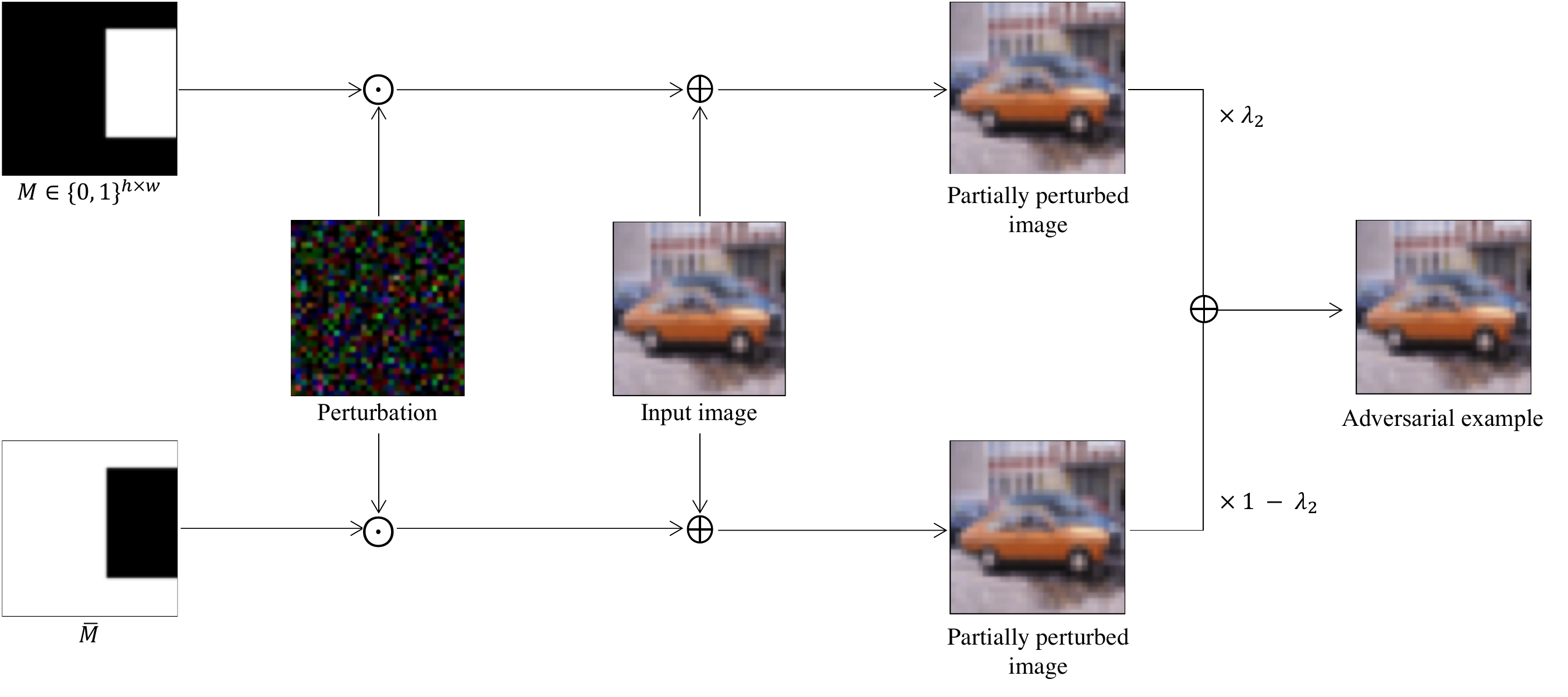}
  \caption{
  Conceptual diagram of computing adversarial examples for \ours .
  Our method separates the perturbation with binary masks, and we mix these two types of adversarial samples stochastically.
  $M$ is a binary mask with a positive value (i.e. 1), for inside the rectangle.
  $\bar{M}$ represents the contrasting binary mask for $M$.
  $\odot$ and $\oplus$ represent element-wise multiply and sum, respectively.
  $\lambda_{2}$ is the interpolation ratio and samples from $\mathrm{Beta}(\alpha, \alpha)$.}
  \label{fig:2}
\end{figure*}

Figure~\ref{fig:2} show the conceptual diagram of \ours, and Algorithm~\ref{alg:1} is the pseudo code of our process.

\begin{algorithm}[tb]
  \caption{Masking and Mixing Adversarial Training}
  \label{alg:1}
  \begin{algorithmic}[1]
  \REQUIRE Training dataset $\mathcal{D}$, batch size $n$, training epochs $T$, learning rate $\eta$, model parameter $\bm{\theta}$, hyper-parameter of beta distribution $\alpha$
  \REQUIRE The function deriving adversarial perturbation $\mathcal{A}$
  \REQUIRE Masking function $\phi$
  \FOR {$t = 1, \dots, T$} 
    \FOR {$\{\bm{x}_{i}, \bm{y}_{i}| i = 1,\dots, n\}\sim \mathcal{D}$}
    \STATE {$\hat{\bm{x}}_{i}\leftarrow \mathcal{A}(\bm{x}_{i}, \bm{y}_{i}; \bm{\theta})$}
    \STATE {$\bm{\delta}_{i}\leftarrow \hat{\bm{x}}_{i} - \bm{x}_{i},~\lambda_{1}\sim U\left[0, 1\right]$}
    \STATE {data masking and label smoothing phase: }
    \STATE {$\bm{\xi}_{i}, \bar{\bm{\xi}}_{i}, \bm{t}_{i}, \bar{\bm{t}}_{i}\leftarrow \phi(\hat{\bm{x}}_{i}, \bm{\delta}_{i},\bm{y}_{i}, \lambda_{1}),~\lambda_{2}\sim \mathrm{Beta}(\alpha, \alpha)$}
    \STATE {data mixing phase: }
    \STATE {$\tilde{\bm{x}}_{i}\leftarrow \lambda_{2}\bm{\xi}_{i} + (1 - \lambda_{2})\bar{\bm{\xi}}_{i}$}
    \STATE {$\tilde{\bm{y}}_{i}\leftarrow \lambda_{2}\bm{t}_{i} + (1 - \lambda_{2})\bar{\bm{t}}_{i}$}
    \STATE {model update: }
    \STATE {$\bm{\theta}_{t+1}\leftarrow \bm{\theta}_{t} - \eta \cdot \frac{1}{n}\sum_{i=1}^{n}\nabla_{\bm{\theta}}\mathcal{L}(f_{\bm{\theta}_{t}}(\tilde{\bm{x}}_{i}), \tilde{\bm{y}}_{i})$}
    \ENDFOR
  \ENDFOR
  \RETURN {model parameter $\bm{\theta}$}
  \end{algorithmic}
\end{algorithm}

\subsection{Masking phase} \label{sec:mask}

In the masking phase, we extract only specific regions of adversarial perturbation computed by Eq.~\ref{eq:pgd} with the binary mask $M\in \{0, 1\}^{h\times w}$, and create partially perturbed images by applying extracted perturbation to clean samples $\bm{x}_{i}$:
\begin{eqnarray}
  \bm{\xi}_{i} &=& \bm{x}_{i} + \bm{\delta}_{i}\odot M,\\
  \bar{\bm{\xi}}_{i} &=& \bm{x}_{i} + \bm{\delta}_{i}\odot (1-M),
\end{eqnarray}
where, $\odot$ is element-wise multiply.
Following to CutMix~\cite{Yun_2019}, we compute the bounding box coordinates $B=(r_{x_{1}},r_{y_{1}},r_{x_{2}},r_{y_{2}})$ for extracting the perturbation with arbitrary probability $\lambda_{1}\sim U[0, 1]$: 
\begin{eqnarray}
  r_{x_{1}}\sim U[0, W],~~r_{x_{2}} = \min \left(W, W\sqrt{1 - \lambda_{1}} + r_{x_{1}}\right),\\
  r_{y_{1}}\sim U[0, H],~~r_{y_{2}} = \min \left(H, H\sqrt{1 - \lambda_{1}} + r_{y_{1}}\right),
\end{eqnarray}
where $H$ and $W$ are respectively height and width of images.
Note, both $r_{x_{1}}$ and $r_{y_{1}}$ sampled from $U[0, W]$ and $U[0, H]$ are not real number $\mathbb{R}$ but integer greater than 0, i.e., $\mathbb{Z}_{\geq0}$.
We eventually get the binary mask $M$ as follows:
\begin{eqnarray}
  M = 
  \begin{cases}
    1 & \text{if}~r_{x_{1}}<M_{:,j}<r_{x_{2}}, r_{y_{1}}<M_{i,:}<r_{y_{2}},\\
    0 & \text{otherwise.}
  \end{cases}
\end{eqnarray}
There is an inverse relation between $\bm{\xi}_{i}$ and $\bar{\bm{\xi}}_{i}$: i.e., with/without the perturbed region is the same.

The ground truth label to perturbed images obtained by these processes does not directly use $\lambda_{1}$; rather, we define smoothed labels with the area ratio between clean and perturbed regions $\lambda_{1}^{\prime}=\frac{\left(r_{x_{2}} - r_{x_{1}}\right)\times \left(r_{y_{2}} - r_{y_{1}}\right)}{H\times W}$:
\begin{eqnarray}
  \bm{t}_{i} &=& \lambda_{1}^{\prime}\bm{y}_{i} + (1 - \lambda_{1}^{\prime})\bar{\bm{y}}_{i}\bm{s},\\
  \bar{\bm{t}}_{i} &=& \lambda_{1}^{\prime}\bar{\bm{y}}_{i}\bm{s} + (1 - \lambda_{1}^{\prime})\bm{y}_{i},
\end{eqnarray}
where $\bm{y}_{i}$ represents a one-hot vector and $\bar{\bm{y}}_{i}$ is all one vector except true class $y_{i}$ (i.e., $1 - \bm{y}_{i}$).
$\bm{s}$ is a uniform distribution assigned uniform probability $1/K-1$ to all class except for true class $y_{i}$.

\subsection{Mixing phase} \label{sec:mix}

AVmixup creates perturbed images by interpolating clean images and adversarial vertex samples with an arbitrary interpolation ratio sampled from $\mathrm{Beta}(1,1)=U[0, 1]$.
However, AVmixup is it limited in terms of the amount of perturbed image variations because they have a uniformly perturbed magnitude over all the images.

Since we attempt to reduce the trade-off by augmenting the perturbed image variation, our method uses the mixing phase for synthesizing two samples computed in the masking phase.
In the mixing phase, we apply the perturbation to the entire image by mixing two attacked images with an arbitrary interpolation ratio, inspired by mixup~\cite{Zhang_2018}.
Note that perturbed images obtained via our method have two types of perturbation magnitude in a single image.

Let $\lambda_{2}\sim \mathrm{Beta}(1, 1)$. 
We then represent mixed sample $\tilde{\bm{x}}_{i}$ and ground truth label $\tilde{\bm{y}}_{i}$ as follows:
\begin{eqnarray}
  \tilde{\bm{x}}_{i} &=& \lambda_{2}\bm{\xi}_{i} + (1 - \lambda_{2})\bar{\bm{\xi}}_{i},\\
  \tilde{\bm{y}}_{i} &=& \lambda_{2}\bm{t}_{i} + (1 - \lambda_{2})\bar{\bm{t}}_{i}.
\end{eqnarray}

This perturbed image is the same as AVmixup with $\gamma=1$ when $\lambda_{1}$ in the masking phase is 0 or 1.
We hope to improve the robustness by enabling the model to train with a richer variation of adversarial examples than AVmixup.

\section{\uppercase{Related work}}

\cite{Szegedy_2014} is the first work that handled the vulnerability of CNNs to malicious noise, called adversarial examples.
Adversarial attack methods for attacking CNNs have since been proposed using various processes.
\cite{Goodfellow_2015}\cite{MD_2016}\cite{Carlini_2017}\cite{Madry_2018} derive adversarial perturbations with the model gradients. 
These attack methods presuppose a situation in which the model parameters are known and are therefore called white-box attacks.
In contrast, the attack methods for situations with unknown model parameters are called black-box attacks.
Papernot~\textit{et al.} is a well-known black-box attack that deals with the transfer-based threat of adversarial examples~\cite{Papernot_2016}.
\cite{MD_2017} proposed universal adversarial examples that do not decide the perturbation for each sample but can induce misclassification of multiple samples with only one perturbation.

A model without countermeasures against adversarial examples can fool even state-of-the-art models.
Adversarial defense studies are being actively conducted to resolve this weakness and various powerful methods have been proposed.
\cite{Goodfellow_2015}\cite{Madry_2018}\cite{Tramer_2018}\cite{Wang_2019}\cite{H.Zhang_2019} guarantee model robustness by updating the model parameters with adversarial examples.
These methods have been known adversarial training.
\cite{N.Papernot_2016} is a defense method with knowledge distillation that makes deriving the gradient difficult by controlling the temperature of the softmax function.
\cite{Meng_2017}\cite{Samangouei_2018} are defensive techniques that prevent performance corruption by removing the perturbation as much as possible from the sample after inputting a classifier.

\section{\uppercase{Experiment}}

We compare our method with several previous works to evaluate its defensive performance.
We use CIFAR-10/-100 as training datasets.
CIFAR-10 is a natural image dataset with ten classes of $32\times 32$ RGB data consisting of 50,000 training samples and 10,000 test samples.
Each class has 5,000 samples for training and 1,000 samples for testing.
CIFAR-100 is almost the same as CIFAR-10 except it has 100 classes, with 500 training samples and 100 test samples in each class.

\subsection{Implementation details}

We use WRN34-10 (Wide Residual Networks)~\cite{Zagoruyko_2016} on both CIFAR-10 and CIFAR-100.
The model trains 200 epochs (80,000 iterations) with a batch size of 128.
The optimizer at training uses momentum SGD with learning rate 0.1, momentum 0.9, and weight decay $2.0\times 10^{-4}$.
The learning rate is a factor of 0.1 at 50\% and 75\% for the number of epochs on CIFAR-10.
Training samples use random crop and horizontal flip as data augmentation and the normalization range [0, 1] for each pixel.
We use the number of rounds $k=10$, epsilon budget $\epsilon=8$, and step size $\alpha=2$ for PGD during our training.
We divide by 255 for each pixel because both $\epsilon$ and $\alpha$ need to fit the training sample scale.

We evaluate the robustness of the trained model with FGSM~\cite{Goodfellow_2015}, PGD~\cite{Madry_2018}, and Carlini \& Wagner (CW)~\cite{Carlini_2017} as adversarial attack methods.
k of PGD-k and CW-k represents the number of rounds for deriving adversarial examples.
We compare our method with the methods below.
\begin{itemize}
  \item Standard: the model trained with standard training data only.
  \item PGD: the PGD-trained model with number of rounds $k=10$ and $\epsilon=8/255$, $\alpha=2/255$.
  \item PGD with LS: the PGD-trained model with label smoothing. 
  A smoothing parameter samples from $\mathrm{Beta}(1, 1)$ during training.
  \item AVmixup: the model trained with the same settings as \cite{Lee_2020}.
\end{itemize}
Experimental results for all methods show the best model, i.e., the highest-robustness model against PGD-20 within training epochs, performance.

\subsection{Comparison results}

\begin{table*}[tb]
\caption{
  Comparison of classification accuracy to clean data and robustness for adversarial attack with various methods.
  PGD with LS represents PGD results using label smoothing with dynamic smoothing parameter.
  Results with * are those referenced from original articles.}
\label{tab:1}
\centering
\begin{tabular}{c|l|ccccc}
\hline
                  Dataset & Model       & Clean & FGSM  & PGD-10 & PGD-20 & CW-20 \\ \hline
\multirow{8}{*}{CIFAR-10} & Standard    & 95.48 & 7.25  & 0.0    & 0.0    & 0.0   \\
                          & PGD         & 85.83 & 58.66 & 52.09  & 50.80  & 30.16 \\
                          & PGD with LS & 86.33 & 61.67 & 55.87  & 54.78  & 30.36 \\
                          & BAT~\cite{Wang_2019}* & 91.2 & 70.7 & -- & 57.5 & 56.2 \\
                          & AVmixup*    & 93.24 & 78.25 & 62.67  & 58.23  & 53.63 \\
                          & AVmixup     & 94.81 & 80.28 & 69.29  & 65.01  & 54.8  \\
                          & \ours~(WRN28-10) & 92.09 & 73.67 & 65.83 & 63.06 & 55.04 \\
                          & \ours       & 93.16 & \textbf{83.35} & \textbf{82.29} & \textbf{80.66} & \textbf{56.90} \\\hline
\multirow{5}{*}{CIFAR-100} & PGD        & 61.29 & 46.01 & -- & 25.17 & -- \\
                           & AVmixup*   & 74.81 & 62.76 & -- & 38.49 & -- \\
                           & AVmixup    & 71.42 & 53.90 & 29.04 & 27.05 & 19.80 \\
                           & \ours      & 69.14 & 32.63 & 32.63 & 29.99 & 18.48 \\\hline
\end{tabular}



\end{table*}

\begin{figure}[tb]
    \centering
    \includegraphics[width=0.98\columnwidth]{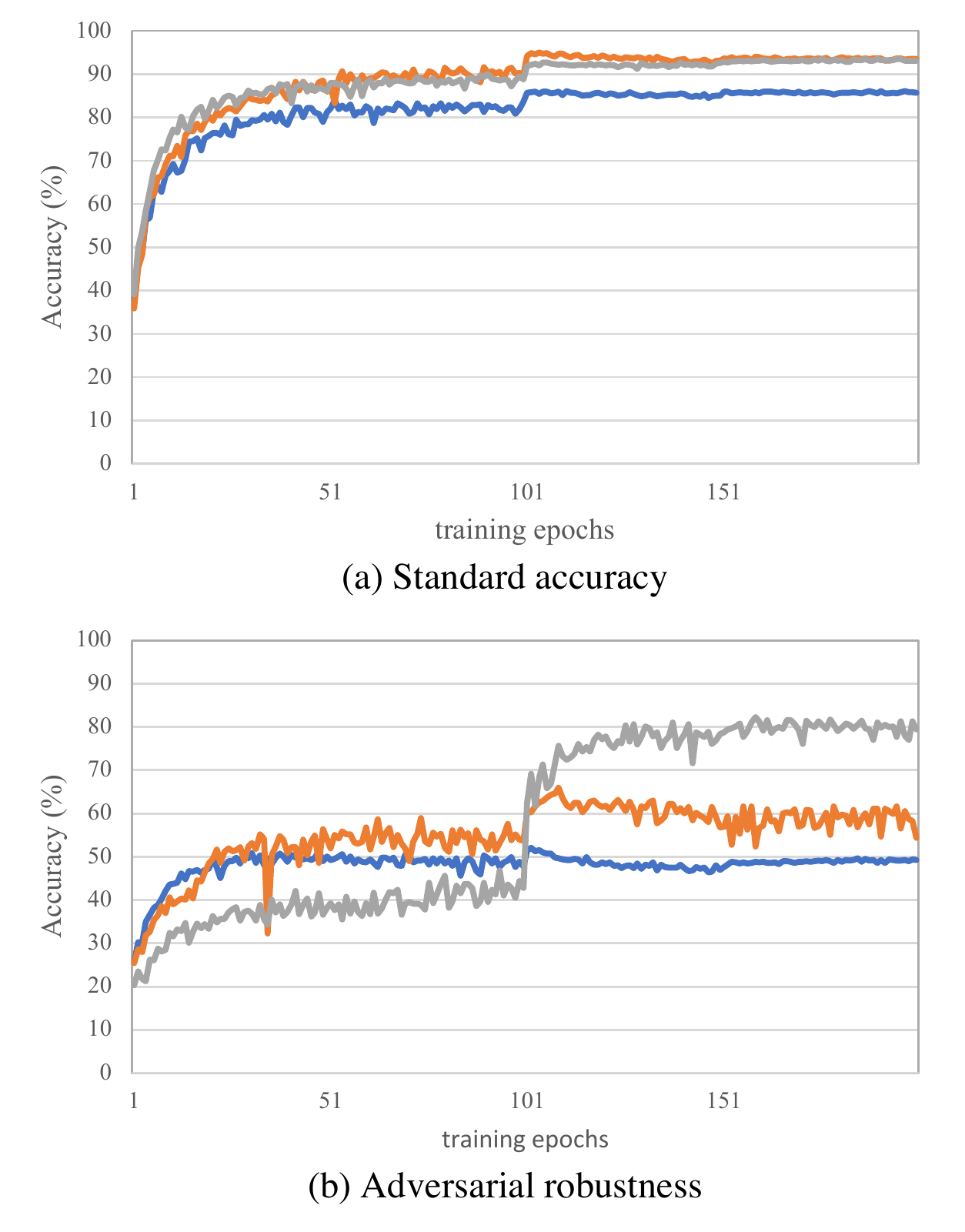}
    \caption{Test accuracy transitions during training of clean data and adversarial examples.
    The perturbation for adversarial example derives by PGD with 10 rounds.
    Blue lines are PGD, orange lines are AVmixup of our implementation, and gray lines are \ours.}
    \label{fig:curve}
\end{figure}

The results on CIFAR-10 are listed in Table~\ref{tab:1}.
\ours~achieved a dramatic performance improvement for the adversarial attack with FGSM and PGD-10/-20.
Its classification accuracy to clean samples was the same as that of AVmixup, and ~\ours~could prevent extreme performance decreases.
For the CW result, classification performance only improved about 3\%p from the AVmixup result.
Both PGD-10 and -20 with our method achived over 80\% and improved by about 30\%p compared to PGD and by about 20\%p compared to AVmixup.
As shown the gray line in Figure~\ref{fig:curve}(b), \ours~could avoid robust overfitting at the same as AVmixup.
The gap between accuracy and robustness dramatically improved for several attacks other than CW by using \ours.
Moreover, the result that fit the base model to BAT~\cite{Wang_2019} outperformed BAT in terms of both clean accuracy and robustness.

In Table~\ref{tab:1}, all methods except ours had a gap of about 10\%p between FGSM and PGD-10/-20 accuracy.
In contrast, our method had hardly any gap between the two.
We discuss this phenomenon in detail in subsection~\ref{subsec:5-4}.

The results on CIFAR-100 in Table~\ref{tab:1} were no better result than the AVmixup results in \cite{Lee_2020}, but our method performed better than our implemented AVmixup.
We could not obtain as good a result as the previous work even when we changed the official implementation on CIFAR-10\footnote{The official implementation of AVmixup: https://github.com/Saehyung-Lee/cifar10\_challenge} to CIFAR-100 and trained.
When we compare the results of PGD and our method, FGSM was comparable and the clean sample and PGD-20 improved by about 7\%p and 10\%p, respectively.
This demonstrates that our method is better able to close the gap between accuracy and robustness than PGD.

Table~\ref{tab:2} shows a performance comparison of our method with TRADES, a defense method with regularization.
While our method could not achieve as good a result as AVmixup based on TRADES, but PGD-20 improved by 10\%p.
Moreover, we could better mitigate the gap between clean accuracy and robustness against perturbed images than AVmixup.

\begin{table}[tb]
  \centering
  \caption{Accuracy comparison with TRADES on CIFAR-10.}
  \label{tab:2}
  \begin{tabular}{c|cc}\hline
                           & Clean & PGD-20 \\\hline
    PGD                    & 87.3  & 47.04 \\
    TRADES (1/$\lambda=1$) & 88.64 & 49.14 \\
    TRADES (1/$\lambda=6$) & 84.92 & 56.61 \\
    AVmixup                & \textbf{90.36} & 58.27 \\
    \ours                  & 89.35 & \textbf{69.76} \\\hline
 \end{tabular}
\end{table}

\begin{table}[tb]
  \centering
  \small
  \caption{Ablation results of our method.}
  \label{tab:3}
  \begin{tabular}{ccc|ccc}\hline
  Masking & Mixing & LS &  Clean & FGSM & PGD-20\\\hline
               &              & $\checkmark$ & 86.33 & 61.67 & 54.78\\
               & $\checkmark$ &              & 89.60 & 54.75 & 44.70\\
               & $\checkmark$ & $\checkmark$ & \textbf{93.97} & 74.81 & 60.96\\
  $\checkmark$ &              &              & 89.92 & 56.36 & 43.72\\
  $\checkmark$ &              & $\checkmark$ & 93.36 & 65.80 & 42.46\\
  $\checkmark$ & $\checkmark$ &              & 90.21 & 60.14 & 49.25\\
  $\checkmark$ & $\checkmark$ & $\checkmark$ & 93.16 & \textbf{83.35} & \textbf{80.66}\\\hline
 \end{tabular}
\end{table}

\subsection{Ablation study}

Table~\ref{tab:3} shows the classification accuracy of our method with combinations ablating every element, such as masking, mixing, and label smoothing.
We use a dynamic value sampled from $\mathrm{Beta}(1, 1)$ as the smoothing parameter.

We can see here that our model achieved the same accuracy as AVmixup using just label smoothing.
The accuracy of PGD-20 was higher than that of AVmixup thanks to creating perturbed images by combined label smoothing and mixing.
This result is equal to training AVmixup with $\gamma=1$ except the smoothing parameter is dynamic.

When training directly adversarial samples with partially applied perturbation using a binary mask, the performance with regards to robustness was inferior to PGD (Table~\ref{tab:1}), but the accuracy to clean samples slightly improved.
This result suggests that we can mitigate the trade-off by absorbing the frequency differences for the clean and adversarial samples with the model.
The same tendency was evident when interpolating adversarial examples and clean samples.

Partially perturbed images further improved the clean accuracy and robustness against FGSM by training using label smoothing, but the PGD-20 result was almost the same.
On the other hand, the accuracy of PGD improved by linearly interpolating two partially perturbed images, but others results, such as clean and FGSM, suffered a decreased performance.
Overall, these results demonstrate that our method is optimal for improving robustness against adversarial examples and maintaining the performance for clean samples.

\subsection{Additional discussion}
\label{subsec:5-4}
\begin{figure*}[tb]
  \centering
  \includegraphics[width=.98\linewidth]{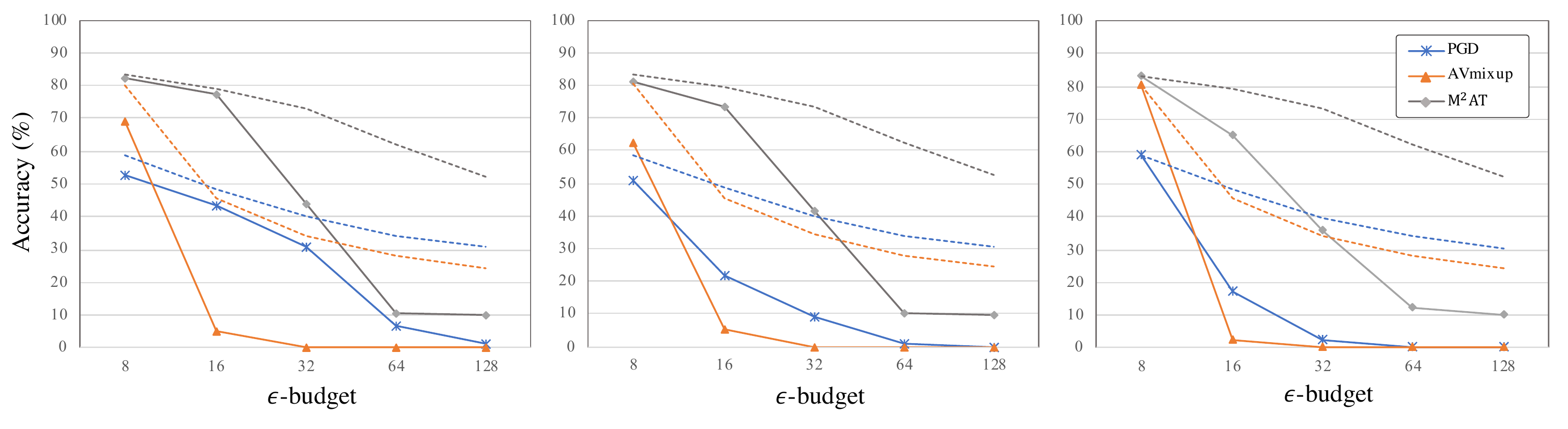}
  \caption{
  Accuracy transitions for adversarial examples at each $\epsilon$ budget.
  From left to right graphs are $\alpha=\{2.0, 4.0, 8.0\}$.
  Solid and dashed lines are PGD-10 and FGSM, respectively, and each method is differentiates by color.
  FGSM accuracy in all graphs was the same.}
  \label{fig:3}
\end{figure*}
\textbf{FGSM vs. PGD for $\mathbf{\mathrm{M^{2}AT}}$}:
Between PGD and FGSM, we found that attacks caused a large gap because the adversarial examples created by PGD provide a more powerful threat than FGSM.
On the other hand, our method had only a small gap between FGSM and PGD (Table~\ref{tab:1}).
To clarify why this is so, we investigate the accuracy transition when expanding the budget size of the adversarial attack.

Figure~\ref{fig:3} shows the accuracy transition for sought perturbations with different budgets and step sizes by FGSM and PGD-10.
Although AVmixup could achieve an excellent performance with $\epsilon$-budget$=8$, its performance dramatically worsened for FGSM and PGD-10 when it explored perturbation with a wide budget over $\epsilon=8$.
This indicates that AVmixup cannot effectively defend against an adversarial attack when the budget size expands to larger than that used during training.
While PGD did not experience so severe a collapse, it is still vulnerable to a large $\epsilon$-budget, the same as AVmixup.
In contrast, we can maintain a superior performance to the other defense methods at all budget sizes.
Moreover, our method maintains a sufficient performance up to $\epsilon$-budget$=32$.
Therefore, the model can handle a budget size much larger than the one used during training if it is trained using our method.
According to these results, AVmixup suffers from an overfit budget size during training and its generalization performance is extremely low, while our method generalizes to a different $\epsilon$-budget.
In other words, we believe that the model with AVmixup has many peak points outside the $\epsilon$-budget.

\textbf{Black-box attack}: 
\begin{table}[t]
  \centering
  \caption{
  Transfer-based black-box attack results (PGD-20).
  The row methods are attack models and the column methods are defense models.
  }
  \label{tab:4}
  \begin{tabular}{l|ccc} \hline
    \multicolumn{1}{c|}{\multirow{2}{*}{\begin{tabular}[c]{@{}c@{}}Defense\\
                        model\end{tabular}}} & \multicolumn{3}{c}{Attack model}\\ \cline{2-4} 
    \multicolumn{1}{c|}{} & \multicolumn{1}{l}{PGD} & \multicolumn{1}{l}{PGD with LS} & \multicolumn{1}{l}{AVmixup} \\ \hline
    PGD          & -- & 50.83 & 50.86 \\
    PGD with LS  & 54.89 & -- & 54.86 \\
    AVmixup      & 64.86 & 64.76 & -- \\
    \ours        & 80.77 & 80.62 & 80.82 \\ \hline
\end{tabular}
\end{table}
We examine the robustness against the transfer-based black-box attack, which attacks the model with perturbations created by other models.
Since the white-box attack is an impossible situation, the black-box attack needs to be robust against perturbation created without knowledge of the parameters of the attack model.
As shown in Table~\ref{tab:4}, our method achieved the best performance against perturbations created by any other models (e.g., PGD, AVmixup).
The demonstrate that the model trained with our method is robust against perturbation created by model gradients other than itself.

\section{\uppercase{Conclusions}} \label{sec:conclusion}

In this paper, we proposed Masking and Mixing Adversarial Training, which can mitigate the trade-off between accuracy and robustness by masking perturbations with a binary mask and mixing them with an arbitrary interpolation ratio.
Experiments showed that \ours~had a better performance than prior works on the CIFAR-10 dataset and was robust against a large $\epsilon$-budget.
While its performance on the CIFAR-100 dataset had the same accuracy as prior works, it could not achieve state-of-the-art performance.
This is because our adversarial examples were slightly different from the original perturbations, which made the model classification is unstable when classes were increased.
We should be able to improve the performance of our method on CIFAR-100 by developing a better method for creating the ground truth label or optimizing the training process.
In future work, we will theoretically investigate our method and improve the training process.

\section*{\uppercase{Acknowledgements}}
This work was supported by JST SPRING, Grant Number JPMJSP2158.

\bibliographystyle{apalike}
{\small
\bibliography{refs}}

\end{document}